\documentclass[letterpaper, 10 pt, conference]{ieeeconf}

\IEEEoverridecommandlockouts                              

\usepackage{graphics} 
\usepackage{epsfig} 
\usepackage{mathptmx} 
\usepackage{times} 
\usepackage{amsmath} 
\usepackage{amssymb}  
\usepackage{subcaption}
\usepackage{mathtools, nccmath}
\usepackage{algorithm,algpseudocode}
\usepackage{hyperref}

\usepackage{booktabs}
\usepackage{multirow}

\usepackage{algorithm}
\usepackage{algpseudocode}
\hypersetup{draft}

\title{\LARGE \bf
Regions of Interest Segmentation from LiDAR Point Cloud for Multirotor Aerial Vehicles
}

\author{Geesara Kulathunga$^{1}$, Roman Fedorenko$^{1}$, Alexandr Klimchik$^{1}$
\thanks{$^{1}$ Center for Technologies in Robotics and Mechatronics Components, Innopolis University, Russia {\tt\small ggeesara@gmail.com, r.fedorenko@innopolis.ru, a.klimchik@innopolis.ru}}%
}

\begin{document}
\maketitle
\thispagestyle{empty}
\pagestyle{empty}

\begin{abstract}

We propose a novel filter for segmenting the regions of interest from LiDAR 3D point cloud for multirotor aerial vehicles. It is specially targeted for real-time applications and works on sparse LiDAR point clouds without preliminary mapping. We use this filter as a crucial component of fast obstacle avoidance system for agriculture drone operating at low altitude. As the first step, each point cloud is transformed into a depth image and then identify places near to the vehicle (local maxima) by locating areas with high pixel densities. Afterwards, we merge the original depth image with identified locations after maximizing intensities of pixels in which local maxima were obtained. Next step is to calculate the range angle image that represents angles between two consecutive laser beams based on the improved depth image. Once the corresponding range angle image is constructed, smoothing is applied to reduce the noise. Finally, we find out connected components within the improved depth image while incorporating smoothed range angle image. This allows separating the regions of interest. The filter has been tested on various simulated environments as well as an actual drone and provides real-time performance. We make our source code, dataset \footnote[2]{Source code and dataset are available at \url{https://github.com/GPrathap/hagen.git}} and real world experiment\footnote[3]{Real-world experiment result can be found on the following link: \url{https://www.youtube.com/watch?v=iHd_ZkhKPjc}}available online.

\end{abstract}

\section{INTRODUCTION}

Over the past two decades, LiDAR technology has been utilized in autonomous vehicles extensively addressing various problems including Multi-Target Tracking (MTT) \cite{choi2013multi}, road and road-edge detection \cite{zhang2010LiDAR}, pedestrian recognition and tracking \cite{wang2017pedestrian}, etc. On the other hand, mass produced Multirotor Aerial Vehicles (MAVs) were not able to apply LiDAR based technologies due to the various limitations such as computational complexity, maximum weight limit and power constrain for longer flight duration. Nevertheless, advancements of the advanced MAVs and faster growth of computational capabilities (e.g., processors, random access memory (RAM), etc.) for executing a large number of instructions per cycle and producing lightweight batteries with high power density, most of the LiDAR technologies and complex algorithms can be executed on board up to some extent. After starting LiDAR as a one of the main sources of visionary sensor, it opens a new paradigm for robust planning and navigation based applications which helps to accomplish different tasks such as surveillance and transportation. The recent work of B. Zhou~\cite{zhou2019robust} and S. Liu~\cite{liu2017planning} propose a quadrotors motion planning system for fast flight in 3-D complex environments only using LiDAR, i.e., visionary sensor.

\begin{figure}[!t]
  \begin{subfigure}[b]{0.45\textwidth}
    \centering
    \includegraphics[width=\linewidth, height=2cm]{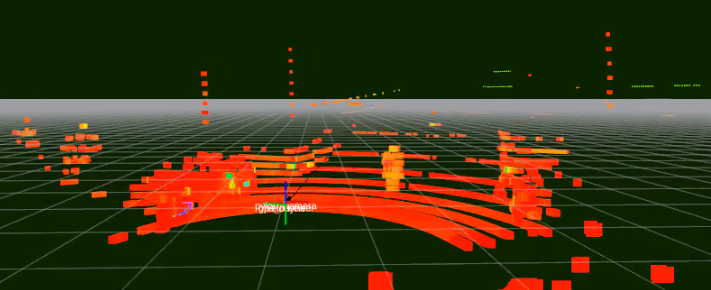}
    \caption{Raw point cloud from Velodyne-16 LiDAR}
    \label{f:row_cloud}
  \end{subfigure}%
  \hfill
   \begin{subfigure}[b]{0.45\textwidth}
    \centering
    \includegraphics[width=\linewidth, height=2cm]{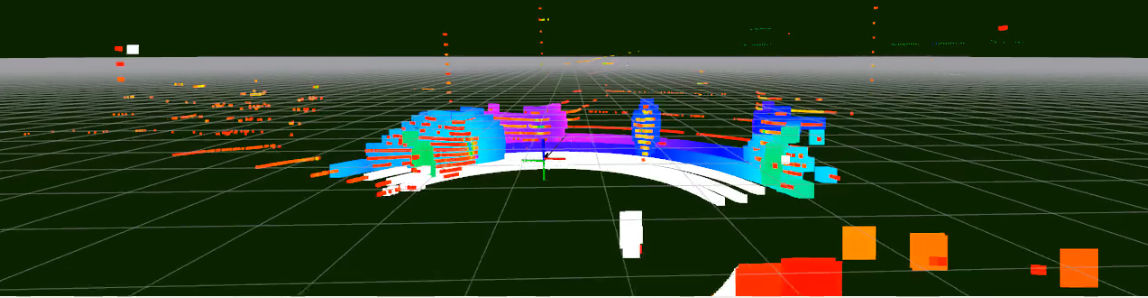}
    \caption{White color represents the regions of uninterest and the other colors depict regions of interest on the point cloud which is shown in Fig~\ref{f:row_cloud}}
     \label{f:separation_on_bfs}
  \end{subfigure}%
  \hfill
  \begin{subfigure}[b]{0.45\textwidth}
    \centering
    \includegraphics[width=\linewidth, height=2cm]{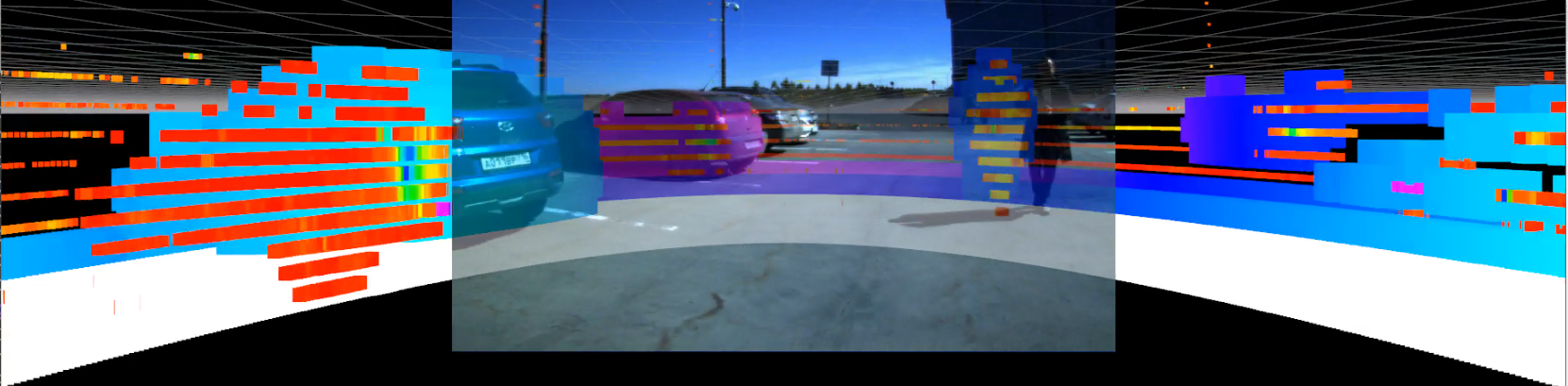}
    \caption{The camera projection on point cloud with the proposed filter result}
     \label{f:projected_on_camera}
  \end{subfigure}
  \caption[Obs Point 1]{The proposed filter results}
  \label{f:main_steps}
\end{figure}

In this work, we propose a new filter called ``Hagen'' for separating regions of interest which specially targeted real-time based applications for MAVs. Thus, it avoids building incremental map that is computationally expensive of the environment. Instead, it uses a set of consecutive point clouds and concatenates them in a sliding window fashion. Afterwards, the concatenated point cloud is utilized for further processing. The proposed filter is designed in a way that it starts detecting objects that are closer to the MAV from the top level to the bottom level of LiDAR laser beams. This feature could use with a path planner, not let MAV to fly into a trap where there is no direction to come out unless it cloud not find the reverse direction. If LiDAR is only the sensor that applied for reasoning of the environment, search space of the LiDAR will eventually become the feasible search space of MAV. Scanning nature of LiDAR helps to start detecting objects from the top to  bottom level. That will help to avoid traversing MAV into terrain or obstacles. How filter processes 3D LiDAR point clouds is shown in Fig.~\ref{f:main_steps}.

\textbf{Our Contributions: }
Proposing a filter to separate regions of interest is the main contribution. Under which, 
\begin{enumerate}
    \item Improving depth image by finding local maxima by applying persistence homology complemented by applying modified bilateral filter for sharpening the detected local maxima
    \item Object detection process starts off detecting objects closer to LiDAR from the top-level (laser beams)
    \item Apply Singular Spectrum Analysis (SSA) for smoothing the range angle image and control the smoothness 
\end{enumerate}

\section{RELATED WORK}

In the work of C.Tongtong\cite{tongtong20113d}, ground segmentation is performed on 3D LiDAR point cloud, building a polar grid map. It apples 1D Gaussian Process (GP) regression model and Incremental Sample Consensus (INSAC) algorithm for segmentation. They have achieved good segmentation
results in the different type of scenarios including urban and countryside environments. The authors of \cite{douillard2011segmentation} present a set of segmentation of dense 3D data algorithms. Empirically they have noticed that prior ground extraction leads to an improvement of the segmentation performance. The work by M. Himmelsbach~\cite{himmelsbach2010fast} shows long-range 3D point clouds segmentation in a real-time and later classification of segmented objects. To reduce the execution time and increase efficiency, their approach was split into solving of two subproblems: ground plane estimation and fast 2D connected components labelling. In our approach, it happens in the opposite direction where fast 2D connected components labelling (regions of interest) is followed by uninterest regions separation. The ground plane also can be within the regions of interest if MAV flies closer to the ground that is the main purpose of doing this way. F. Moosmann~\cite{moosmann2009segmentation} presents a fast algorithm that works with a high volume of 3D LiDAR data in a real-time. It uses a novel unified generic criterion based on local convexity measures for separation of ground and non-ground which is based on a graph data structure. The authors of \cite{Sabirova2019} present an implementation of the ground detection methodology with filtration of forest points from LiDAR-based dense 3D point cloud using the Cloth Simulation Filtering (CSF) algorithm. The methodology requires dense mapping and offline analysis but allows to recover a terrestrial relief and create a landscape map of a forestry region. 

Most of the 3D point cloud segmentation techniques over the last decade were based on an either statistical or probabilistic model. But B. Wu~\cite{wu2018squeezeseg} proposes a novel approach for semantic segmentation of road-objects from 3D LiDAR point clouds based on neural networks. They have formulated the problem as a point-wise classification problem using a convolutional neural network (CNN). A transformed LiDAR point cloud is taken as the input for the model. Point-wise label map or cluster will be the output of the model. The recent development of hardware helps to run such high computing power demanding algorithms effectively. The authors of ~\cite{maturana20153d}, also uses 3D CNN for detecting of small and potentially obscured obstacles in vegetated terrain incorporating volumetric occupancy map.  Initially, they need to train a model what sort of terrain to be detected from raw occupancy data. 

\begin{figure*}[!t]
\begin{center}
\includegraphics[width=18cm, height=6cm ]{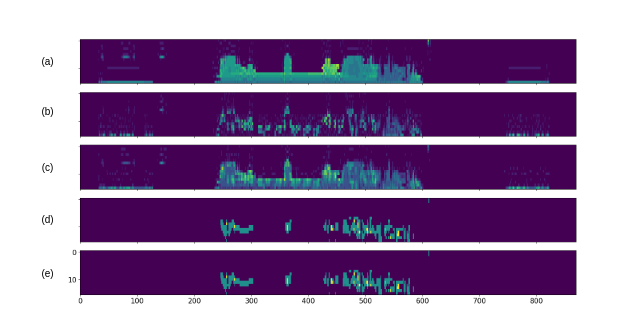}
\caption{\label{f:main_steps_of_hagen} (a) depicts the depth image ($depth\_img$) which is transformed from a point cloud (corresponds to point cloud shown in Fig.~\ref{f:row_cloud}). The result of local maxima detection ($filtered\_img$) is shown in (b). Mean image ($processed\_img$) is shown in (c). Range angle image $angle\_img$ and smoothed range angled image ($smoothed\_angle\_img$) are shown in (d) and (e) respectively}
\end{center}
\end{figure*}

Filtering is one of the most fundamental techniques of computer vision. For example, a Gaussian filter computes the weighted average around the given pixel location with a given size of a kernel. Bilateral filter also uses a somewhat similar approach but preserving edges around each considered pixel. In this work\cite{premebida2016high}, formulate obtaining dense depth-map using local spatial interpolation that depends on sliding window-based approach in which BF is employed. They have modified conventional BF to achieve proper upsampling which can preserve foreground-background discontinuities. In that way they were able to acquire high resolution depth-maps by upsampling of 3D-LiDAR data. In this research, Bilateral Filter (BF) is applied to sharpen locations around local maxima detected as shown in Fig.~\ref{f:main_steps_of_hagen}.

\begin{figure}[!t]
\begin{center}
\includegraphics[width=\linewidth, height=2.6cm]{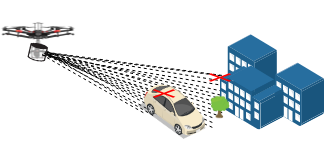}
\caption{\label{f:detection_idea} Illustration of the superposition of objects between close, i.e., part of the car and far, i.e., tree and part of the building whose distance or depth from LiDAR will be in the same neighborhood region of the depth image. Nearest points in the point cloud from LiDAR are marked with the red crosses}
\end{center}
\end{figure}

Usually, appropriate smoothing is applied on the filtered output to improve the robustness. The main intuition of smoothing a signal is to construct an approximate signal which tries to capture meaningful patterns while leaving out noise or other rapid phenomena. SSA is one of the smoothing techniques which gives a highly accurate and concise result. SSA is utilized extensively in the biomedical related application for smoothing signals and find some hidden patterns within it (e.g., \cite{pataky2019smoothing, mourad2019ecg} and \cite{prathap2018near}).

\section{Separating regions of interest} \label{ground_non_ground_separation}

Typically, ground plane can be separated simply removing points that are lower than the location of the LiDAR (assume that installed location is known) or RANSAC-based plane fitting. However, none of those methods does work for MAVs because MAV altitude can change anytime in contrast to ground vehicles. Hence, ground plane also can be considered within the regions of interest if it flies closer to the ground. The main steps of the proposed filter is provided in Algorithm.~\ref{al:overview}. Inputs for the filter: $point\_cloud$, $ground\_removal\_angle$ (explain in Sec. ~\ref{sec:construction_vertical_angles}), $number\_of\_principle\_components$ and $smoothing\_window\_size$ (explain in Sec. ~\ref{sec:ssa_smoothing}).

\begin{algorithm}
\caption{Separation of regions of interest}
\label{al:overview}
\textbf{Input} $point\_cloud$, $number\_of\_principle\_components$ (PC), $ground\_removal\_angle$ ($\beta$), $smoothing\_window\_size$ (W)\\
\textbf{Output} $region\_of\_interest$, $non\_region\_of\_interest$ 
\begin{algorithmic}[1]
\State $depth\_img \gets ConstructDepthImage()$
\State  $filtered\_img \gets ApplyLocalMaxima2DFinder()$
\State $processed\_depth\_img \gets (depth\_img + filtered\_img)/2$
\State $angle\_img \gets ConstructAngleImage()$
\State $smoothed\_angle\_img \gets ApplySSASmoothing()$
\State $ roi, non\_roi \gets LabelRegionOfInterest()$
\end{algorithmic}
\end{algorithm}


\subsection{Depth Image Estimation from Point Cloud} \label{s:depth_estimation}

We used Velodyne VLP16 LiDAR for our experiments, however the proposed algorithm could be used with other LiDARs (e.g, Velodyne HDL-64E, HDL-32E, etc). VLP16 provides the point cloud which consists of 16 laser beams that covers $\pm$15 degrees vertically and 360 degrees horizontally. Processing point cloud itself is time-consuming which is not suited for real-time analysis. Thus, we have decided to reduce space dimension 3D to 2D where disparity map or depth map is employed for separation of the regions of interest. Width of the depth image is set for 870 ($I_w$). 870 is selected empirically under which horizontal resolution becomes 0.413($360/870$) degrees/pixel. It can be changed, but it may affect the performance and the accuracy. Depth image height $(I_h)$ is set for 16 because VLP16 LiDAR has only 16 channels. To transform a point cloud into a depth image, each point in the point cloud to be assigned to a pixel in the depth image. Distance or depth to each considered point corresponds to an intensity of the chosen pixel. If there are multiple points corresponded to the same pixel location, we keep the closest distance from the LiDAR. Once non-ground is separated in 2D space, it is needed to acquire the corresponding points in the point cloud from the depth image. Thus, while estimating the depth image, we keep a record which points belongs to which pixel location. This mapper is named the $depth\_mapper$ which will be using in the following sections.

Following steps are required to create the depth image: 
\begin{enumerate}
    \item Let ($x_i, y_i, z_i : i=1,..., N$) are the points in a cloud,  where $N$ is the number of points 
    \item Estimate angles on $x$ and $y$ directions for a given point: 
        \begin{equation}
         \begin{aligned}
        angle\_x_i = asin(\frac{z_i}{dist_i}),  \
      angle\_y_i = atan2(y_i, x_i) 
        \end{aligned}   
    \end{equation}
    
    \item 360 degrees are split into $I_w$ number of pixels; 30 degrees are split into $I_h$ number of pixels. Then we get closest matching row and column that corresponds with $angle\_x$ and $angle\_y$ in degrees. It will be the pixel location on the depth image. Here $dist_i = \sqrt{x_i^2 + y_i^2 + z_i^2}$ is the intensity value. 
\end{enumerate} 
A similar idea has proposed in \cite{bogoslavskyi2017efficient} as well. After performing this, depth image can be constructed as shown in Fig.~\ref{f:main_steps_of_hagen}a. 

\subsection{Local Maxima Detection in Depth Image}
Intuition of detecting local maxima locations is shown in Fig.~\ref{f:detection_idea}. We are interested in finding locations of high pixel intensity (local maxima) which correspond to closest objects from current pose of MAV. Extracting local maxima from given images that are in high-dimensional space is generally challenging because of incompleteness and noise. 
In applied mathematics, Topological Data Analysis (TDA)\cite{chazal2017introduction} is an approach for analyzing  data by using topological properties. This allows to find some hidden structures or patterns. Persistence Homology (PH)\cite{weinberger2011persistent} is one of the dominant mathematical tools employed for this purpose.
PH provides a proper way of analysing such data while being less sensitive to a particular metric and transforming original space into different space which is robust to noise. 

\begin{figure}[!t]
\begin{center}
\includegraphics[width=\linewidth, height=3cm]{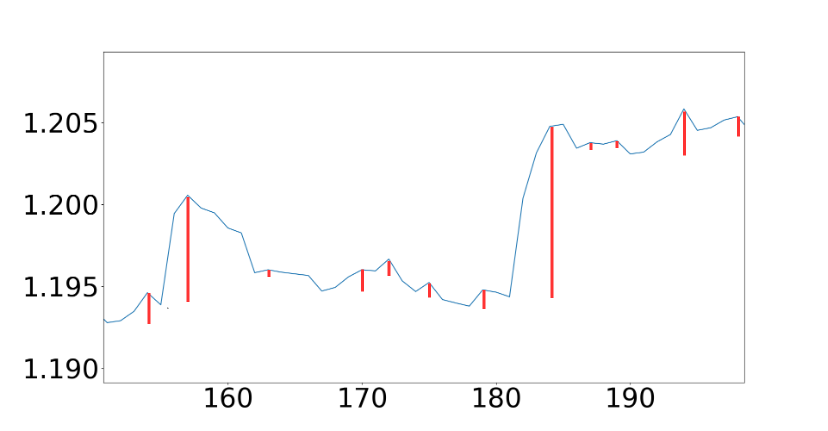}
\caption{\label{f:persistent_map}  Blue color curve shows how normalized pixel intensities are propagated over given row of image. Red color vertical lines represent barcode for the most significant local minima}
\end{center}
\end{figure}

\begin{figure}[!ht]
\begin{center}
\includegraphics[width=\linewidth, height=3cm]{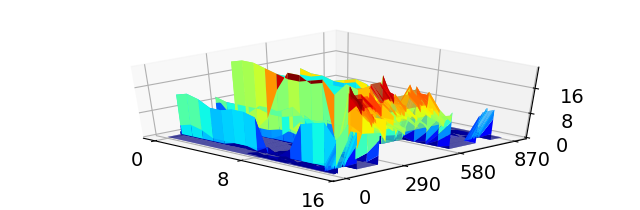}
\caption{\label{f:local_minima_detation} Local minima detection of $depth\_img$ where blue color plane depicts the pixel locations of $depth\_img$ and strength of local mamixa is varied from blue to red along the orthogonal direction to the image plane}
\end{center}
\end{figure}

To get a clear intuition of how PH applies for local maxima detection, let's consider Fig. \ref{f:persistent_map}. Though we are interested in local maxima, some of the local maxima might be irrelevant for time to time. For example, local maxima near 155 must be considered while discarding local maxima at 165. Those false positives are arisen because of the noise of the image. Hence, false positives should be eliminated. Noise can be removed by smoothing the image. Nonetheless, this comes with a cost (e.g., some of sharpen local maxima get vanished or dampened, intensity level of pixel will change if there is no proper normalization, etc). Thus, it would be better to operate on the original data itself. In Persistent Homology, there is a concept called barcode. A barcode represents each persistent generator with a vertical line which begins at the first filtration level when it arises and ending at the second filtration level when it disappears. Persistent generator, first filtration level and second filtration level are corresponded to distance between detected local maxima to consecutive local minima, local maxima and local minima respectively. In Fig.~\ref{f:persistent_map}, red colored lines depict the most persistent 0-dimensional cycles~\cite{kurlin2015one} where the local maxima have been detected. Hence, PH is employed for detecting relative maxima in noise-resilient fashion in 2D space as shown in Fig.~\ref{f:local_minima_detation}.

Hence, PH finds the relative maxima in noise-resilient fashion.  Once the most persistent local maxima have been detected, it is needed to sharpen those locations in depth image without affecting neighbour pixels. A bilateral filter (BF) is one of the techniques to achieve this because BF is a non-linear, edge-preserving, and noise-reducing smoothing filter. It replaces the intensity of each of the pixels with a weighted average of intensity values from nearby pixels. We have modified BF slightly different way which get more accurate result compare to the original BF~\cite{tomasi1998bilateral}. The modified BF is given as follows.

\begin{equation}
    I^{filtered}(x) = \frac{1}{W_p} \Sigma_{x_i \in \Omega} I(x)G(x-x_i, \sigma_x)G(dis(x, x_i), \sigma_n) + I(x)
\end{equation}
and normalization term, $W_p$ is defined as 
\begin{equation}
    W_p = \Sigma_{x_i \in \Omega} G(x-x_i, \sigma_x)G(dis(x, x_i), \sigma_n)
\end{equation}

\begin{equation}
   G(\mu, \sigma) = \frac{1}{\sigma \sqrt{2\pi}}\exp^{\frac{-1}{2}(\mu/\sigma)^2}, \quad dis(x, x_i) = \sqrt{x^2+x_i^2} 
\end{equation} where $I^{filtered}(x)$ is the filtered image, I is the depth image, x are the coordinates of local maxima that are detected, $\Omega$ is the 8 neighbours around each x, $G(\mu, \sigma)$ is a Gaussian function where $\mu$ is the mean and $\sigma$ is the standard deviation, $\sigma_x$ and $\sigma_n$ are the standard deviations which are constant all the time; values are set as 1.2 and 1.3 respectively which were estimated empirically. $dis(x, x_i)$ is utilized to estimate distance between x and $x_i \in \Omega$ where $i = 1,...,8$.

\subsection{Constructing Range Angle Image} \label{sec:construction_vertical_angles}
\begin{figure}[!ht]
\begin{center}
\includegraphics[width=\linewidth,height=4cm]{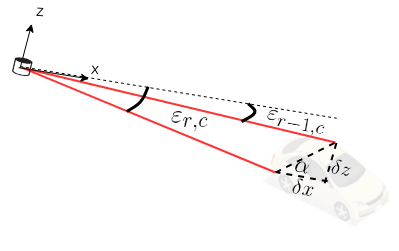}
\caption{ Illustration of how $\alpha$\label{f:angle_estimation} is calculated. To construct range angle image, $(I_h-1)I_w$ number of $\alpha$ angles to be calculated as follows}
\end{center}
\end{figure}

In each iteration, the depth image is followed by range angle image are constructed. Since laser beams cover $\pm$15 degrees vertically and $I_h$ is 16, it is needed to calculated 15 vertical angles in which each angle estimated as angle between the horizontal plane of the LiDAR and considered laser beam which varies from LiDAR to LiDAR. As shown in Fig.~\ref{f:angle_estimation}, $\varepsilon_{r,c}$ and $\varepsilon_{r-1,c}$ are two vertical angles that correspond to two consecutive rows with a considered column, i.e., r-1,c and r,c. Thus, each range angle ($\alpha$) of the range angle image is calculated considering depths (i.e., depicts in red color lines in Fig.~\ref{f:angle_estimation}) at considered consecutive rows and a column in depth image. Hence, the range angle image is constructed with 15x870 ($(I_h-1)I_w$) resolution. Let define two consecutive laser beams that are projected on a car which are shown in red colored lines in Fig.~\ref{f:angle_estimation}. Since $\varepsilon_{r,c}$ and $\varepsilon_{r-1,c}$ are known, $\alpha$ can be calculated as follows

\begin{equation}
     \begin{aligned}
    \alpha  = atan2(\delta z, \delta x) , \\
    \delta z = |D_{r-1,c} sin(\varepsilon_{r-1,c}) - D_{r,c}sin(\varepsilon_{r,c})|, \\
    \delta x = |D_{r-1,c} cos(\varepsilon_{r-1,c}) - D_{r,c} cos(\varepsilon_{r,c})| 
    \end{aligned}   
\end{equation} where $D_{r,c}$ is distance or depth at (r,c) in the depth image. As shown in Fig. ~\ref{f:angle_estimation}, $\varepsilon_{r,c}$ and $\varepsilon_{r-1,c}$ are vertical angles in between r and r+1 rows which corresponds with $c$-th column. $(r, c) = (1, ..., I_h-1; 0,.., I_w-1)$

\subsection{Smoothing Range Angles with SSA} \label{sec:ssa_smoothing}

SSA\cite{hassani2007singular} is a powerful non-parametric spectral estimation technique for time series analysing and forecasting. SSA incorporates elements of classical time series analysis, multivariate statistics, dynamical systems and signal processing. In this work, SSA has been chosen over other smoothing techniques because of several reasons. When MAV flies closer to ground, ground plane should be considered, if not only objects closer to MAV should be considered. SSA provides this capability together with depth image based on the smoothing factor ($PC$). Following section explains how it is being achieved.  

SSA is applied for smoothing out the range angle image in column-wise. Let define angles in each column, $\Theta = (\alpha_0, ..., \alpha_{N-1}) \in \mathbb{R}$ of length N ($I_h$). The SSA algorithm consists of two main steps: decomposition of a given series and reconstruction by adding desired principal components. 

\subsubsection{Decomposition}

In this stage, $\Theta$ is rearranged into WxK matrix which is called as the trajectory matrix $(X \in \mathbb{R})$. W is the window length or $smoothing\_window\_size$ $(1 < W < N)$. K can be formulated as  $K = N - W + 1$. We fix the value for W as 8 even it is a configurable parameter for the filter. This assumption is due to each column consist of 16 pixels and two iterations were assumed to be enough that made empirically. Furthermore, if you are going to use 32 or 64 channel LiDAR it is better to increase $W$. 
\begin{equation}
    X = [X_1 : ... : X_K] , \qquad  X_j = [\theta_{j-1}, ..., \theta_{j+W-2}]^t
\end{equation} where $j = 1, ..., K$. Then Singular Value Decomposition (SVD) is applied on X, where jth component of SVD is specified by jth eigenvalue $\lambda_j$ and eigenvector $U_j$ of $XX^t$. 

\begin{equation}
    X = \Sigma^d_{j=1} \sqrt{\lambda_j}U_jV_j^t, \qquad V_j =  \frac{X^tU_j}{\sqrt{\lambda_j}} 
\end{equation} where $d = max \{j: \lambda_j > 0\}$. Since X is a Hankel matrix (or catalecticant matrix) and $XX^t$ is a positive-define matrix,  their eigenvalues ($\lambda_j$) are positive as well. Then, eigenvectors of X are ordered in decreasing order of corresponding eigenvalues. 

\subsubsection{Reconstruction}

\begin{figure}[!ht]
\begin{center}
\includegraphics[width=\linewidth, height=4cm]{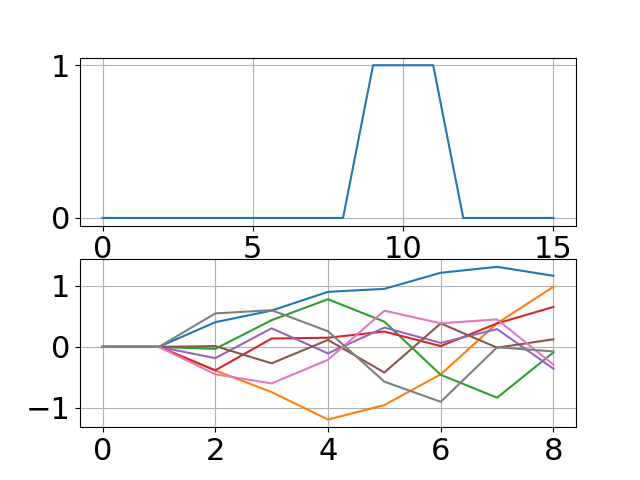}
\caption{ Principal components extraction corresponds to 370th column of Fig.~\ref{f:main_steps}d. In this figure, the first subfigure depicts the intensity changes over column-wise. The second subfigure represents corresponding principal components \label{f:principle_components}}
\end{center}
\end{figure}

In this stage, select a set of SVD components ($PC$) and averaging along entries (hankelization) with indices of the X from the selected components of the SVD. More about hankelization can be found in~\cite{golyandina2001analysis}. In Fig.~\ref{f:principle_components}, it is shown how principal components are extracted. Based on $PC$, smoothness will vary. If all the principal components are taken, it implies that there is no smoothing present under which ground plane can be present within the interest of regions. Thus, PC should be selected based on the area where MAV flies.  

\subsection{Connected Components Extraction with BFS} \label{sec:connected_components}

\begin{algorithm}[!ht]
\caption{Connected components extraction with BFS}
\label{alg:connected_components}
\begin{algorithmic}[1]
\Procedure{SeparationRegionofInterest()}{}
 \State $selected\_depth\_points \gets []$
 \State $labeled_image \gets []$
 \State $selected\_depth\_val \gets Inf$
  \For{c  = 0, ..., $I_w$}
   \State $selected\_depth \gets -1, -1$
   \State $depth\_is\_set \gets false$
    \For{r  = $I_h$-1, ...,0}
            \State $depth \gets processed\_depth\_img(r, c)$
            \If{$depth > 0.01 \& !depth\_is\_set$}
              \State $depth\_is\_set \gets true$
              \State $selected\_depth\_points \gets \; push (r, c) $
            \EndIf
            \If{$depth < selected\_depth$}
              \State $selected\_depth\_val \gets depth$
              \State $selected\_depth \gets \; (r, c) $
            \EndIf
    \EndFor
    \If{$selected\_depth\_val > 0.01$}
              \State $selected\_depth\_points \gets \; push selected\_depth $
    \EndIf
\EndFor
\For{$depth\_point : selected\_depth\_points$}
       \State LabelConnectedComponent($depth\_point$, $ground\_removal\_angle, labeled\_image$)
\EndFor
\EndProcedure
\end{algorithmic}
\end{algorithm}
This is the last step of the proposed filter. In Algorithm.~\ref{alg:connected_components}, it is given the procedure of the process and $LabelConnectedComponent()$ is utilized for finding connected components using the breadth-first search (BFS). The idea used was initially proposed in \cite{bogoslavskyi2016fast}, a few modifications have been made to improve performance and accuracy. Once regions of interest are labelled, corresponding cloud points can be extracted using the $depth\_mapper$ (refer to the Sec.~\ref{s:depth_estimation}). The result of this is shown in Fig.~\ref{f:separation_on_bfs}.  

\section{EXPERIMENTAL EVALUATION}

This section mainly focuses on evaluating specific aspects of the proposed filer detailed in Sec.~\ref{ground_non_ground_separation} and assessing the performance of the system. Since the filter is designed as a parametric model, there are three hyperparameters: PC, $\beta$ and W are needed to fine tune for getting a stable result. To evaluate how those affect on the final result, initially LiDAR and filter output (point clouds) are projected on camera as shown in Fig.~\ref{f:projected_on_camera} within the camera's field of view (FOV). Thus, evaluation is done only on the region covered by camera's FOV considering as an image segmentation problem. Hence, $F_1$ is used as the evaluation metric which is one of the well-known widely used evaluation matrices in image segmentation applications.

\begin{equation}
     \begin{aligned}
    F_1 = 2\frac{P*R}{P+R}, \      
   P = \frac{t_p}{t_p+f_p}, \
   R =\frac{t_p}{t_p+f_n} \\
    \end{aligned}   
\end{equation} where $P$ denotes the precision, $R$ is for recall, $t_p, t_f, f_p$ and $f_n$ are true positive, true negative, false positive and false negative respectively. The $F_1$ score is defined as the harmonic mean of P and R. In mathematics, harmonic is the most suitable tool for work with rates and ratios. Since precision and recall are ratios, it is one of the most appropriate tools for estimating the filter accuracy. 
 \begin{figure}[!h]
  \centering
  \begin{subfigure}[b]{0.45\textwidth}
    \centering
    \includegraphics[width=\linewidth, height=1.5cm]{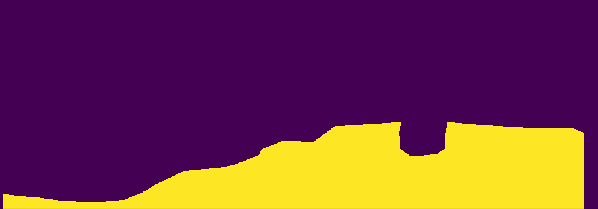}
    \caption{Ground truth (manually labeled) region of uninterest }
    \label{f:ground}
  \end{subfigure}%
  \hfill
   \begin{subfigure}[b]{0.45\textwidth}
    \centering
    \includegraphics[width=\linewidth, height=1.5cm]{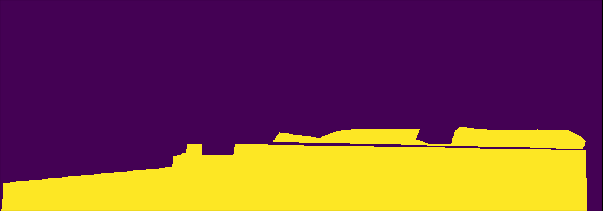}
    \caption{Filter output (region of uninterest only) after the camera projection}
     \label{f:labeled_ground}
  \end{subfigure}%
  \hfill
  \begin{subfigure}[b]{0.45\textwidth}
    \centering
    \includegraphics[width=\linewidth, height=1.5cm]{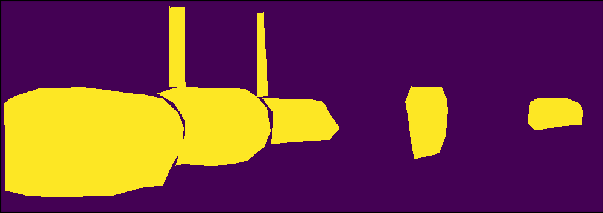}
    \caption{Manually labeled regions of interest}
    \label{f:non_ground}
  \end{subfigure}%
  \hfill
   \begin{subfigure}[b]{0.45\textwidth}
    \centering
    \includegraphics[width=\linewidth, height=1.5cm]{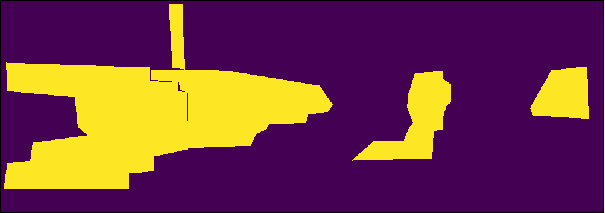}
    \caption{Filter output (regions of interest only) after the camera projection}
     \label{f:labeled_non_ground}
  \end{subfigure}%
  \caption{Labeled images correspond with Fig.~\ref{f:main_steps}}
\end{figure}
 As the first part of the evaluation, a set of images consisting of projected LiDAR point cloud within camera's FOV was manually selected. And we segmented them into two separate classes: regions of interest and region of uninterest as shown in Fig.~\ref{f:ground} and Fig.~\ref{f:non_ground}. Afterword, we projected filter output on camera space as regions of interest and region of uninterest separately and labelled them as shown in Fig.~\ref{f:labeled_ground} and Fig.~\ref{f:labeled_non_ground}. In this way, 50 images were labelled for the each of the classes, overall 200 images for a given specific parameter configuration (PC, $\beta$ and W). As mentioned in Sec.~\ref{sec:ssa_smoothing}, W is fix for this experiment. Initially, we ran the proposed filter on a set of ROS bags several times while changing PC and $\beta$ values. Along with that, we came up with a hypothesis for the proper value combination of the hyperparameters, i.e., PC (5) and $\beta$ (10 degrees). To claim our hypothesis was correct or wrong, PC and $\beta$ were varied around the observed values. Table~\ref{t:beta_angle_estimation} and Table~\ref{t:pincipal_components} show how those affect the final result of the proposed filter.

\begin{figure}[!ht]
\begin{center}
\includegraphics[width=\linewidth, height=3.5cm]{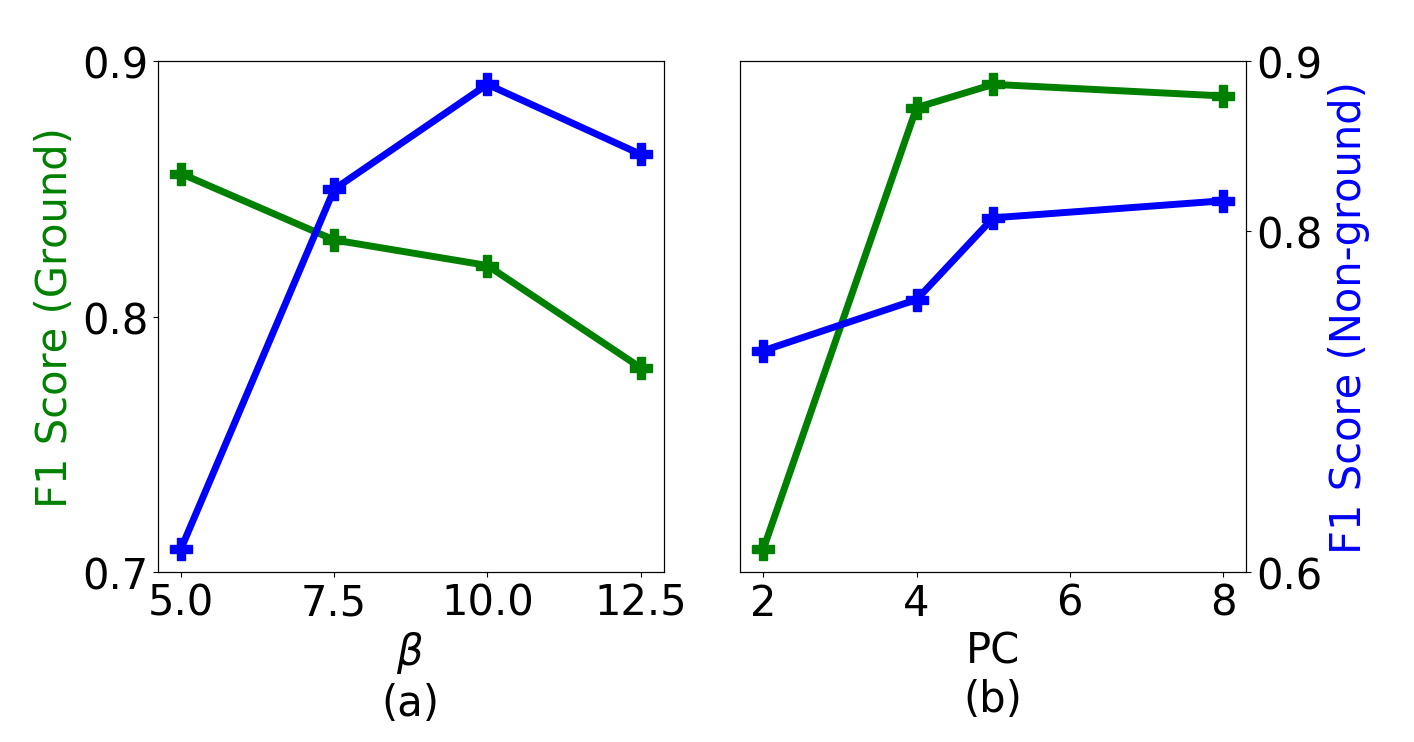}
\caption{ The accuracy of proposed filter computed as $F_1$ score for separation of uninterest regions (Table.~\ref{t:beta_angle_estimation}) and interest regions (Table.~\ref{t:pincipal_components}). $\beta$ angle (degree) and PC are the hyperparameters that are varied one at a time while holding the other hyperparameters constant}
\label{f:f1_and_angle}
\end{center}
\end{figure}
 
 According to the Fig.~\ref{f:f1_and_angle}a, $F_1$ score of regions of interest is steeply increased from degree of 5 to peak of 10 degree. After that, it has a downwards trend. On the other hand, there is a dip in $F_1$ for region of uninterest. After 10 degrees, both $F_1$ scores have a decline. As shown in Fig.~\ref{f:f1_and_angle}b,  $F_1$ of region of uninterest is sharply increased from 2 to peak of 5. After that it has a steadiness. It is common for the $F_1$ of regions of interest as well with less score. Still, there is space to increase PC value to achieve high $F_1$ score. We have decided not to increase PC further because it may be a overestimation of the accuracy of the filter. 
 
 To assess the performance of the system, we ran the filter through 3000 point clouds as 3 mini batches (1000 per each).  Then, we measured average running time of the proposed filter and its standard deviation per 360 degrees Velodyne-16. Result is given in Table~\ref{t:performace}.

\begin{table}[t]
  \centering
  \caption{Changing the angle $\beta$ while fixing PC (5) to evaluate how $\beta$ will affect on the final accuracy of the filter}
   \begin{tabular}{|c|c|c|c|c|c|c|}
    \hline
   \multirow{2}{*}{Angle $\beta$} & \multicolumn{3}{c|}{Region of Uninterest} & \multicolumn{3}{c|}{regions of interest} \\
    & P & R & F1 & P & R & F1\\
    \hline
    \multirow{1}{*}{5} &0.948&0.77&0.856&0.825&0.63&0.715\\
    \hline
    \multirow{1}{*}{7.5} &0.931&0.748&0.83&0.872&0.717&0.787\\
    \hline
     \multirow{1}{*}{\textbf{10}} &\textbf{0.925}&\textbf{0.762}&\textbf{0.82}&\textbf{0.892}&\textbf{0.738}&\textbf{0.808}\\
    \hline
     \multirow{1}{*}{12.5} &0.897&0.7&0.78&0.862&0.735&0.794\\
    \hline
  \end{tabular}
  \label{t:beta_angle_estimation}
\end{table}

\begin{table}[t]
  \centering
  \caption{Changing the PC while fixing $\beta$ (10 degree) to evaluate how PC will affect  on the final accuracy of the filter }
   \begin{tabular}{|c|c|c|c|c|c|c|}
    \hline
   \multirow{2}{*}{PC} & \multicolumn{3}{c|}{Region of Uninterest} & \multicolumn{3}{c|}{regions of interest} \\
    & P & R & F1 & P & R & F1\\
    \hline
    \multirow{1}{*}{2} & 0.715&0.547&0.62&0.79&0.678&0.73\\
    \hline
    \multirow{1}{*}{4} &0.917&0.7253&0.81&0.82&0.708&0.760\\
    \hline
    \multirow{1}{*}{\textbf{5}} &\textbf{0.925}&\textbf{0.762}&\textbf{0.82}&\textbf{0.892}&\textbf{0.738}&\textbf{0.808}\\
    \hline
    \multirow{1}{*}{8} &0.908&0.739&0.815&0.905&0.746&0.818\\
    \hline
  \end{tabular}
  \label{t:pincipal_components}
\end{table}

\begin{table}[th!]
 \centering
 \caption{Average running time and its standard deviation per 360 degrees LiDAR scan}
\begin{tabular}{|l|l|}
\hline
\begin{tabular}[c]{@{}l@{}}Nvidia Xavier\\ ARMv8.2 @ 2.2GHz\end{tabular} & \begin{tabular}[c]{@{}l@{}}Acer F5-573g \\ i5-6200U @2.30 GHz\end{tabular} \\ \hline
34 ms $\pm$ 12 ms  &  25 ms $\pm$ 15 ms 
\\ \hline
\end{tabular}
\label{t:performace}
\end{table}

\begin{figure}[!ht]
\begin{center}
\includegraphics[width=\linewidth, height=4cm]{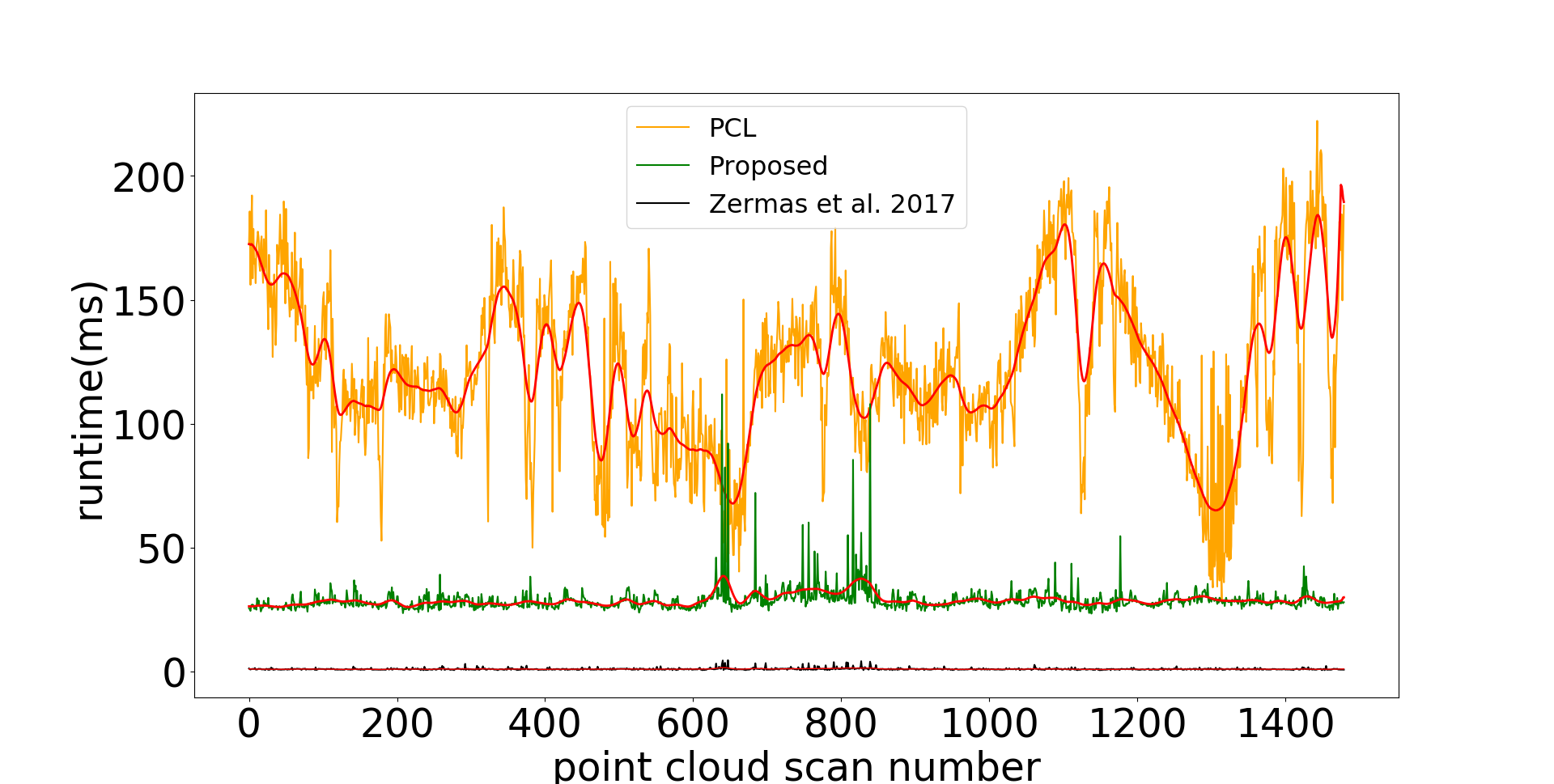}
\caption{Execution time approximately for 1,500 point clouds from a 16-beam Velodyne dataset with the proposed approach, Zermas's~\cite{zermas2017fast} approach and euclidean segmentation from PCL~\cite{rusu20113d}}
\label{f:speed_test}
\end{center}
\end{figure}

In order to compare accuracy of the proposed filter with existing methods, we have selected two other methods: fast segmentation of 3D point clouds~\cite{zermas2017fast} and euclidean segmentation from PCL~\cite{rusu20113d}. When selecting existing methods we have not considered neural network based methods. Those are not able to run on real MAVs due to low GPU capabilities. 1500 point clouds scans were processed by each of the filters. The result of the speed test is shown in Fig.~\ref{f:speed_test}. Runtime for euclidean segmentation from PCL was considerably higher than the proposed filter. When comparing proposed filter with Zermas~\cite{zermas2017fast} proposed idea, the proposed filter quite shower. This is because of finding local maxima takes more than the 50\% of the total execution time per iteration.  

\begin{figure}[!ht]
\begin{center}
\includegraphics[width=\linewidth, height=3.5cm]{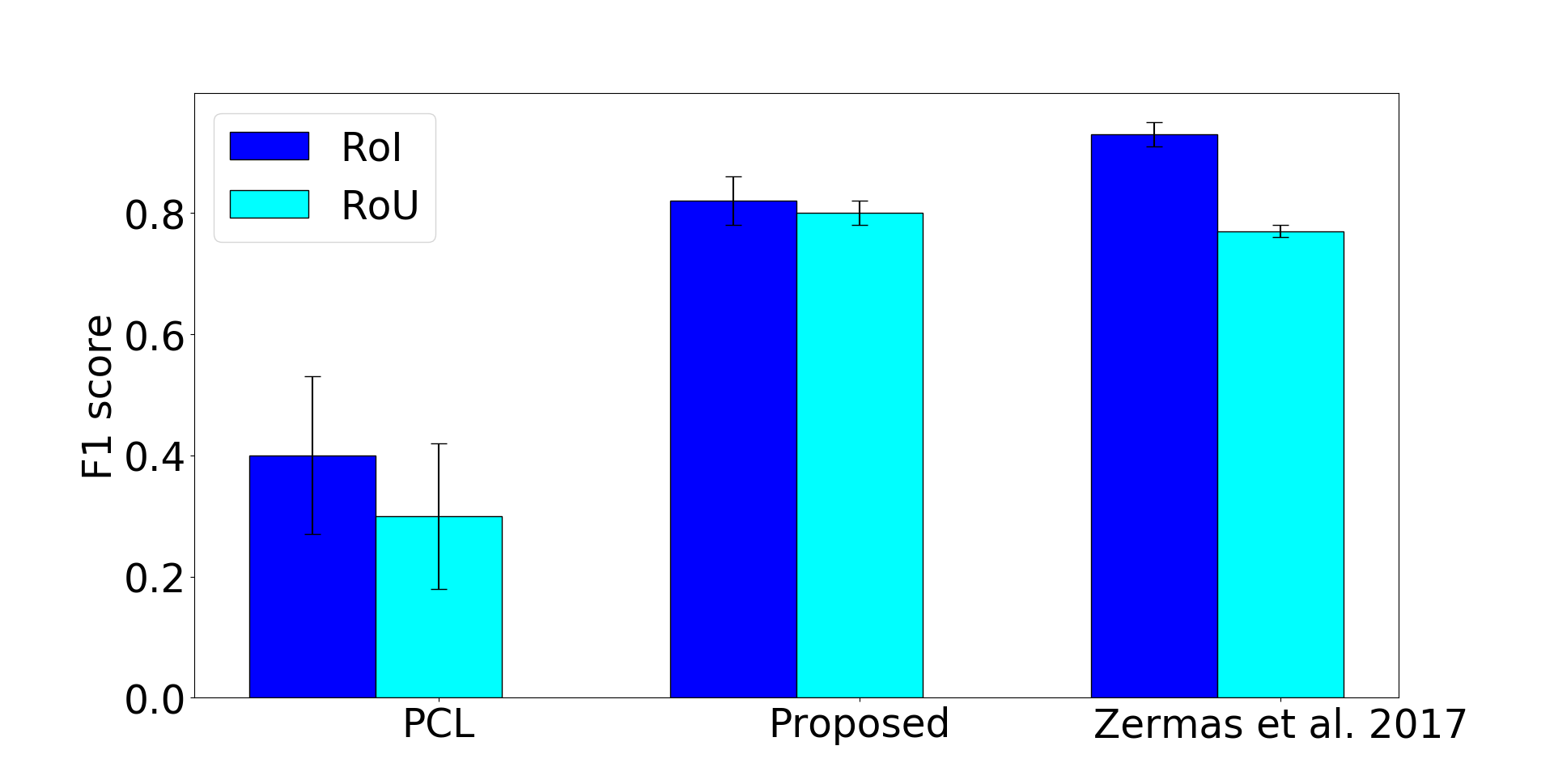}
\caption{RoI and ROU stand for regions of interest and region of uninterest respectively. F1 score of RoI quite higher for Zermas~\cite{zermas2017fast} compared to the proposed filter. On the other hand, area not interested in has slightly higher F1 score for the proposed filter}
\label{f:precesion}
\end{center}
\end{figure}

Finally, we have compared the accuracy of the proposed filter using F1 score. To calculate F1 score for the Zermas's~\cite{zermas2017fast} method and euclidean segmentation from PCL method, same approach which is used to find the accuracy of the proposed solution has been utilized with the same dataset. Result are shown in Fig.~\ref{f:precesion}. F1 score of RoI quite higher for Zermas~\cite{zermas2017fast} compared to the other two methods. On the contrary, region of uninterest has slightly higher F1 score for the proposed filter compared to Zermas's~\cite{zermas2017fast} method. In addition, we have more flexibility how to adjust the filter parameters based on the environment conditions where the MAV flies.

\section{CONCLUSIONS}
In this paper, we have presented a complete filter for segmenting regions of interest on LiDAR point cloud. The system was designed and implemented with a focus on MAVs real-time applications. Filter works on sparse LiDAR point clouds without preliminary mapping.
In particular, we have presented main steps of the proposed algorithm – identity locations of high densities (local maxima) within depth image, merging of  original depth image with identified locations after maximizing intensities around local maxima and utilizing range angle image to search connected components in the improved depth image for segmenting out regions of interest.

Finally, we validated our approach on simulated and real conditions with a series of experiments. The accuracy and performance of the proposed filter is compared with competitive existing methods. 

\section*{ACKNOWLEDGMENT}
The work presented in the paper has been supported by Innopolis University, Ministry of Education and National Technological Initiative in the frame of creation Center for Technologies in Robotics and Mechatronics Components (ISC 0000000007518P240002)

\bibliographystyle{IEEEtran}
\bibliography{IEEEabrv,IEEEexample}

\end{document}